\documentclass[10pt,twocolumn,letterpaper]{article}

\usepackage[pagenumbers]{wacv} 

\definecolor{wacvblue}{rgb}{0.21,0.49,0.74}
\usepackage[pagebackref,breaklinks,colorlinks,allcolors=wacvblue]{hyperref}

\usepackage{microtype}
\usepackage{multirow}
\usepackage[table]{xcolor}
\usepackage{makecell}
\usepackage{graphicx}
\usepackage{subcaption}
\usepackage{booktabs} 
\definecolor{goldenbrown}{RGB}{245, 222, 179}
\definecolor{lightgreen}{RGB}{220,240,220}
\usepackage{wasysym} 
\newcommand{\on}{\CIRCLE}
\newcommand{\off}{\Circle}
\usepackage{pifont}
\newcommand{\cmark}{\ding{51}} 
\newcommand{\xmark}
{\ding{55}} 

\def\wacvPaperID{703} 
\def\confName{WACV}
\def\confYear{2027}

\title{CountZES: Counting via Zero-Shot Exemplar Discovery and Selection}

\author{
Muhammad Ibraheem Siddiqui \qquad
Muhammad Haris Khan \\
Mohamed Bin Zayed University of Artificial Intelligence \\
{\tt\small \{muhammad.siddiqui, muhammad.haris\}@mbzuai.ac.ae}
}

\begin{document}
\maketitle
\begin{abstract}
Zero-shot object counting (ZOC) aims to count instances of unseen categories using only a class name. Existing exemplar-based methods discover candidate exemplars through random image patches or open-vocabulary detector proposals. However, random patch sampling often yields poorly delineated instances, while detector-based discovery may produce multi-instance regions and, in dense scenes, provide only a few or even no suitable single-instance candidates. Thus, exemplar selection is constrained by candidate quality and coverage, while counting itself remains highly sensitive to exemplar choice. To this end, we propose CountZES, an inference-only framework that jointly addresses candidate discovery and exemplar selection through density-guided candidate generation and complementary exemplar-selection mechanisms. Starting from a text-grounded anchor, CountZES progressively discovers and refines visual exemplars through three stages: Detection-Anchored Exemplar (DAE), Density-Guided Exemplar (DGE), and Feature-Consensus Exemplar (FCE). DAE establishes a reliable single-instance anchor, DGE leverages density responses to expand candidate discovery beyond the detector-constrained pool and select a statistically consistent exemplar, and FCE promotes representativeness through feature-level consensus. Together, they yield complementary exemplars balancing semantic grounding, count consistency, and feature representativeness. CountZES requires no ZOC-specific optimization and adopts a modular design that supports different pretrained components. Extensive evaluations across natural, aerial, and medical domains demonstrate strong cross-domain generalization and superior performance among inference-only ZOC methods. Code is available at \href{https://github.com/Muhammad-Ibraheem-Siddiqui/CountZES}{GitHub}.
\end{abstract}

\section{Introduction}
\label{sec:intro}

\begin{figure}[t]
\centering
\includegraphics[width=1\linewidth]{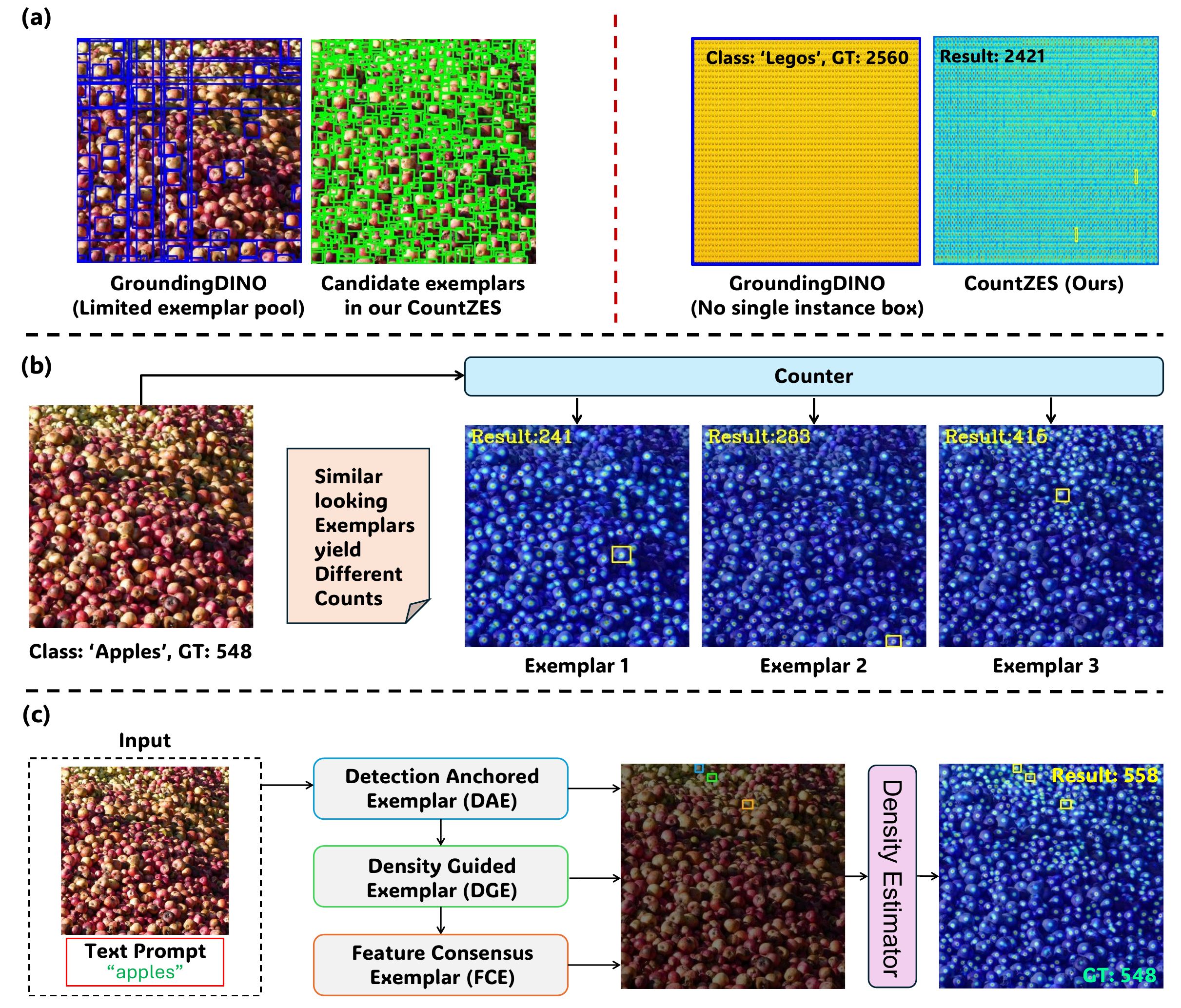}
\caption{\textbf{Motivation for CountZES.} (a) Detector-driven exemplar selection is constrained by the initial candidate pool, as GroundingDINO~\cite{liu2024grounding} may provide few or no clean single-instance proposals in dense scenes. (b) Counting is sensitive to exemplar choice, with visually similar exemplars yielding different count estimates. (c) CountZES addresses both challenges through progressive discovery and selection, yielding complementary exemplars from semantic, count-driven, and feature cues. (Best viewed in zoom).}
\label{fig:intro}
\vspace{-5pt}
\end{figure}

Object counting is traditionally \textit{class-specific}, tailored for predefined categories such as crowds~\cite{babu2022completely}, vehicles~\cite{li2020bi}, or cells~\cite{tyagi2023degpr}. To improve scalability, \textit{class-agnostic counting} aims to generalize to unseen categories~\cite{yang2021class,huang2023interactive,hui2024class}, spanning \textit{few-shot object counting} (FSOC), \textit{reference-less counting} (RLC), and \textit{zero-shot object counting} (ZOC). FSOC requires annotated exemplars at inference~\cite{zhai2024zero,zhang2025enhancing}, while RLC relies on global visual cues~\citep{hobley2022learning,d2024afreeca} but lacks semantic specificity~\cite{xu2023zero,zhu2024zero}. ZOC instead specifies arbitrary unseen categories through text prompts alone~\citep{jiang2023clip,xu2023zero,kang2024vlcounter}. Existing ZOC methods broadly follow two paradigms. The first learns text-conditioned representations for counting: CLIP-Count~\cite{jiang2023clip} and VLCounter~\cite{kang2024vlcounter} align textual and visual representations for density regression, while CountSE~\cite{liu2025countse} learns soft exemplars for text-guided counting. A complementary paradigm, which we focus on, discovers visual exemplars directly from the query image and uses them to guide an exemplar-based counter. This separation enables a modular design where the text prompt identifies what to count from the query image, while counting leverages an existing pretrained exemplar-based density estimator without ZOC-specific optimization. Existing visual-exemplar approaches differ primarily in how candidate exemplars are obtained. ZSOC~\cite{xu2023zero} selects class-relevant exemplars from arbitrarily sampled image patches, which frequently contain partial objects, background, or poorly delineated instances. VA-Count~\cite{zhu2024zero} instead uses GroundingDINO~\cite{liu2024grounding} proposals followed by single-object filtering, but remains restricted to detector-proposed candidates, which may contain multiple instances or provide few suitable single-instance regions in dense scenes, Fig.~\ref{fig:intro}(a). This exposes a fundamental limitation: \textit{exemplar selection is constrained by the quality and coverage of its candidate pool}. Random sampling provides coverage but inaccurate delineation, whereas detector-based discovery provides stronger localization but a restricted candidate space. Moreover, valid candidates are not equally suitable for counting. Visually similar single-instance exemplars can yield substantially different density estimates, Fig.~\ref{fig:intro}(b). Therefore, reliable zero-shot exemplar discovery requires both a sufficiently rich candidate pool and a principled mechanism for selecting representative exemplars.

\begin{figure}[t]
\centering
\includegraphics[width=1\linewidth]{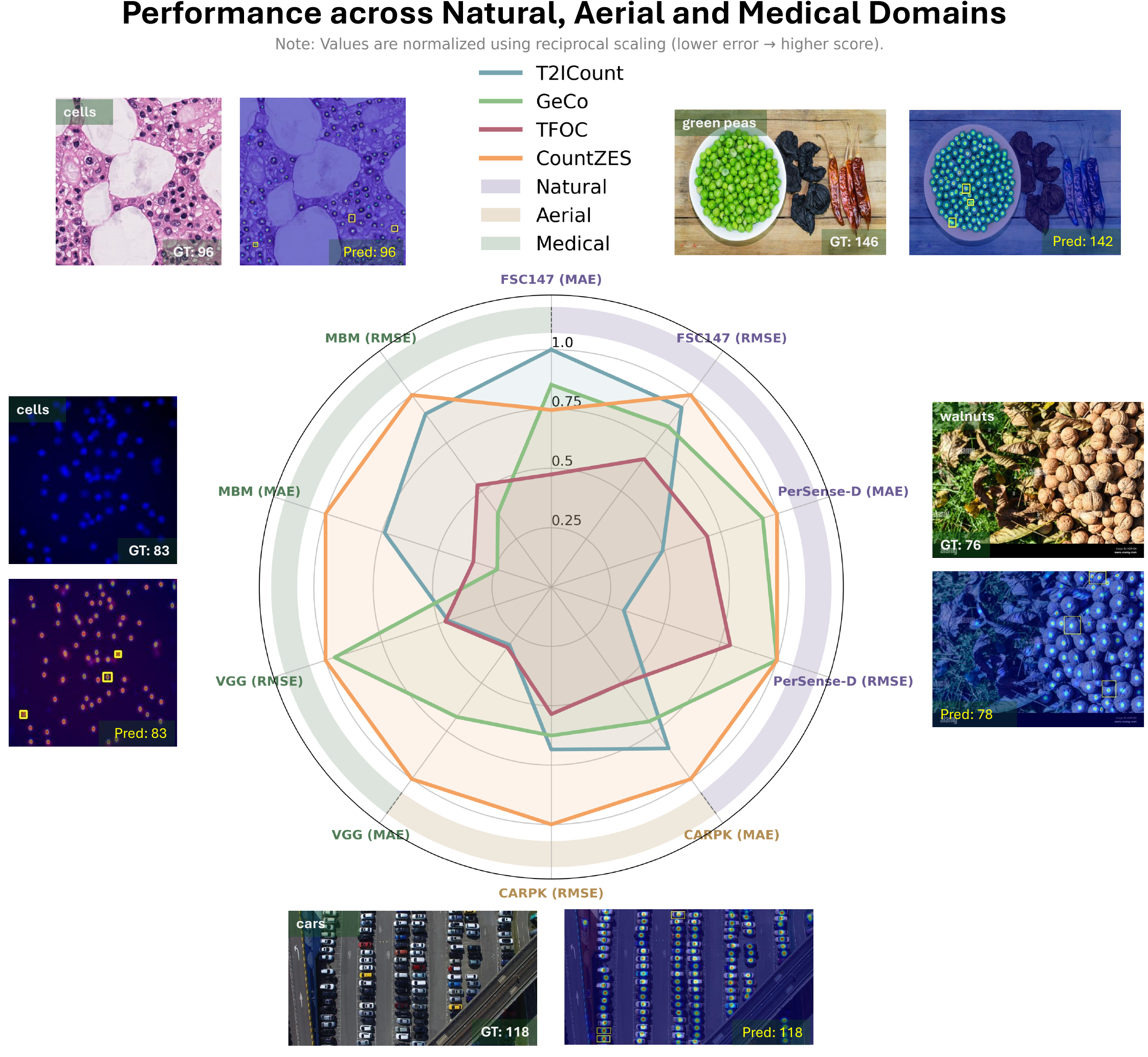}
\caption{CountZES comparison with T2ICount~\cite{qian2025t2icount}, GeCo~\cite{pelhan2024novel}, and TFOC~\cite{shi2024training} across natural, aerial, and medical domains.}
\label{fig:teaser}
\vspace{-5pt}
\end{figure}

Although density estimation is particularly effective in dense scenes, where detection-based counting becomes less reliable~\cite{liu2018decidenet}, its potential for zero-shot \textit{exemplar discovery} remains largely underexplored. We address this by reconsidering the role of density estimation. Rather than using a density estimator only for the final count, we exploit its spatial response to discover instances resembling an initial text-grounded exemplar. This creates a progressive mechanism where an exemplar guides density estimation, the density response expands candidate discovery beyond the initial detector proposals, and the resulting candidates enable more reliable exemplar selection. Our goal is to retain density estimation as the counting mechanism without learning a new text-conditioned counter or performing ZOC-specific optimization, as ZOC-trained approaches such as T2ICount~\cite{qian2025t2icount}, which explicitly optimize text-conditioned counting, can exhibit limited transfer under substantial domain shifts (Fig.~\ref{fig:teaser}). Inference-only methods avoid such optimization, but largely formulate counting as a segmentation task. TFOC~\cite{shi2024training} and TFCAC~\cite{lin2025simple} rely solely on SAM~\cite{kirillov2023segment}, which lacks semantic awareness~\cite{ma2023can}. SAM3~\cite{carion2025sam} introduces text-prompted segmentation, but still struggles to separate small, densely packed instances, as reflected by its performance on FSC-147~\cite{ranjan2021learning} benchmark (see Sec.~\ref{experiments}). This motivates an unexplored direction: using a pretrained exemplar-based density estimator for both zero-shot exemplar discovery and counting, without ZOC-specific optimization.


To this end, we introduce CountZES, an inference-only framework for zero-shot visual exemplar discovery and selection, Fig.~\ref{fig:intro}(c). CountZES operates progressively through three stages. \textit{Detection-Anchored Exemplar (DAE)} establishes a text-grounded single-instance anchor from coarse detector proposals. Conditioned on this anchor, \textit{Density-Guided Exemplar (DGE)} exploits the density response to discover candidates beyond the detector-constrained pool and selects a statistically consistent exemplar. Finally, \textit{Feature-Consensus Exemplar (FCE)} promotes visual representativeness through feature-level consensus. Together, the stages yield complementary exemplars balancing textual alignment, count consistency, and feature representativeness. CountZES performs no ZOC-specific optimization or fine-tuning; its pretrained components remain fixed and are not jointly optimized with the downstream counter. This modular design enables strong cross-domain generalization across natural, aerial, and medical domains (Fig.~\ref{fig:teaser}). Our contributions are:

\begin{itemize}

\item We propose CountZES, an inference-only ZOC framework that moves beyond random sampling or detector-constrained candidate pools by exploiting density responses to progressively discover visual exemplars.

\item We introduce a three-stage exemplar selection framework comprising DAE, DGE, and FCE. DAE establishes a reliable text-grounded single-instance anchor, DGE leverages density responses to discover a richer candidate pool and select a statistically consistent exemplar, and FCE promotes visual representativeness through feature consensus.
\item CountZES requires no ZOC-specific optimization and adopts a modular design that supports different pretrained components. Extensive experiments across natural, aerial, and medical domains demonstrate strong cross-domain generalization, robustness to imperfect detector proposals, and the benefit of cross-stage exemplar collaboration.

\end{itemize}

\section{Related Work}
\label{sec:relatedwork}

To eliminate the need for manually annotated exemplars in FSOC~\cite{lu2018class,ranjan2021learning,liu2022countr,djukic2023low,he2024few}, Xu \textit{et al.}~\cite{xu2023zero} introduce ZOC, where objects are counted from a class name alone. Existing methods broadly follow learned text-conditioned counting, visual-exemplar discovery, or inference-only segmentation paradigms. Image--text approaches~\cite{jiang2023clip,kang2024vlcounter,amini2023open} learn cross-modal representations for density estimation, while T2ICount~\cite{qian2025t2icount} exploits diffusion priors and CountSE~\cite{liu2025countse} learns soft exemplar representations. These methods require ZOC-specific optimization and can exhibit limited transfer under domain shift (Fig.~\ref{fig:teaser}). Closer to our work are methods that explicitly discover visual exemplars from the query image. ZSOC~\cite{xu2023zero} selects exemplars from randomly sampled patches using class prototypes and predicted counting errors, but such patches may contain partial objects or background. VA-Count~\cite{zhu2024zero} instead filters GroundingDINO~\cite{liu2024grounding} proposals for single-object exemplars, but remains restricted to detector-proposed candidates, which can include multi-instance regions or provide few suitable instances in dense scenes.  CountZES differs by using an initial text-grounded exemplar to obtain density responses that enable candidate discovery beyond the detector-constrained pool, followed by statistical and feature-level selection of complementary exemplars. Other approaches integrate detection or segmentation more directly. DAVE~\cite{pelhan2024dave} adopts a detect-and-verify paradigm, PseCo~\cite{huang2024point} combines point prediction, SAM~\cite{kirillov2023segment}, and CLIP~\cite{radford2021learning}, while GeCo~\cite{pelhan2024novel} jointly models detection, segmentation, and counting but requires task-specific supervision. In contrast, inference-only methods such as TFOC~\cite{shi2024training}, TFCAC~\cite{lin2025simple}, and OmniCount~\cite{mondal2025omnicount} formulate counting as a segmentation task and inherit SAM's limitations in dense scenes~\cite{ma2023can}. SAM3~\cite{carion2025sam} introduces promptable concept segmentation, but struggles in dense scenes. CountZES instead retains exemplar-driven density estimation, proven effective for dense counting~\cite{liu2018decidenet}, while performing zero-shot visual exemplar discovery.

\section{Method}
\label{sec:method}

\begin{figure*}[t]
    \centering
    \includegraphics[width=0.95\linewidth]{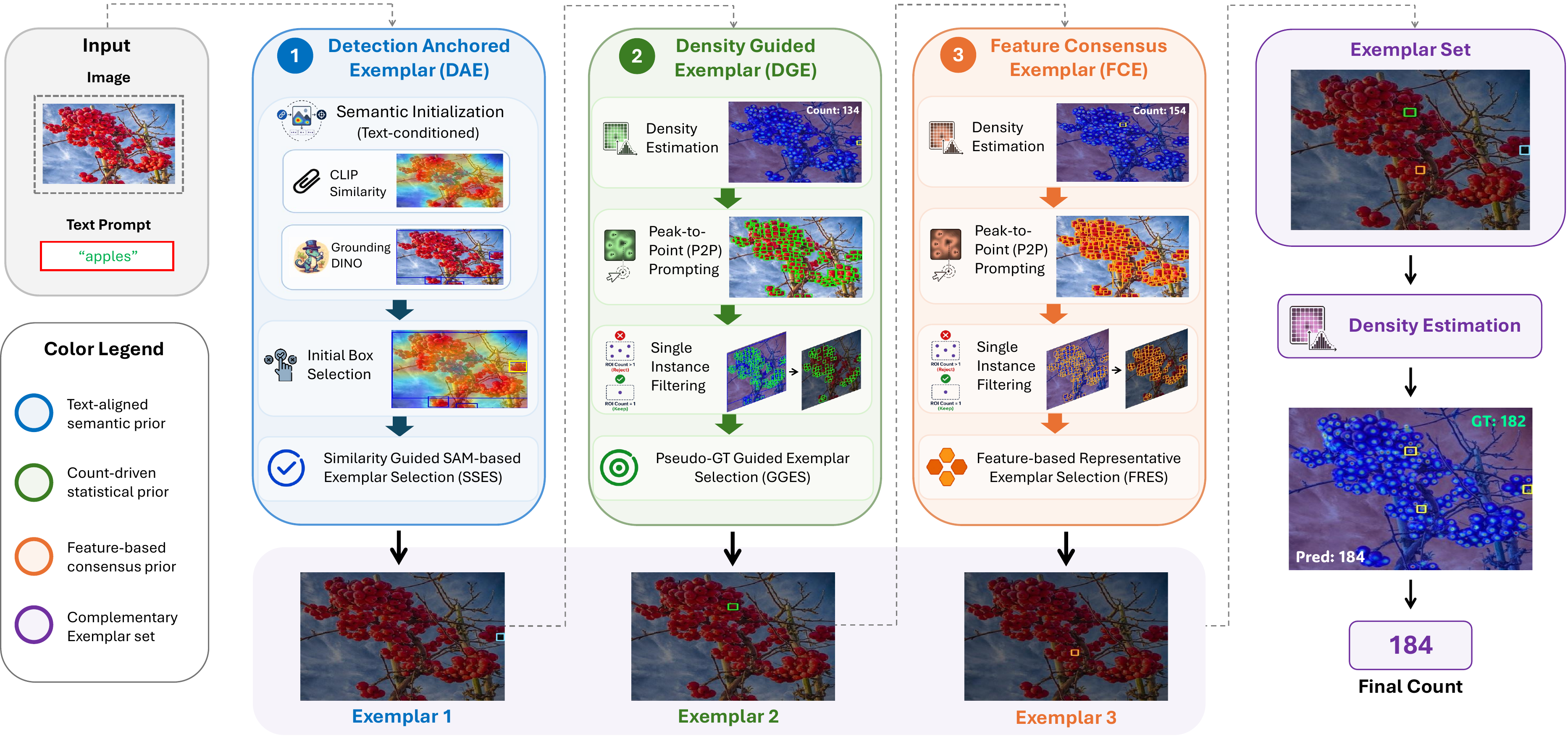}
  \caption{\textbf{Overview of CountZES.} Given an image and text prompt, CountZES progressively discovers and selects visual exemplars through three stages. DAE refines GroundingDINO proposals using CLIP similarity and SSES to obtain a text-aligned single-instance anchor. Conditioned on this anchor, DGE leverages the density response and P2P prompting to discover candidate instances, applies RoI-based single-instance filtering, and selects a statistically consistent exemplar through GGES. FCE further promotes visual representativeness by projecting density-guided candidates onto SAM's feature space and selecting a representative exemplar through FRES. The exemplars selected across the three stages form a complementary set for final density estimation and counting.}
    \label{fig:mainfig}
    \vspace{-4pt}
\end{figure*}


CountZES performs zero-shot exemplar selection through three synergistic stages (Fig.~\ref{fig:mainfig}). DAE (Sec.~\ref{DAE}) establishes a semantically grounded single-instance anchor, DGE (Sec.~\ref{DGE}) uses density responses for candidate discovery and statistically consistent selection, and FCE (Sec.~\ref{FCE}) promotes visual representativeness via feature-level consensus.




\subsection{Detection Anchored Exemplar (DAE)}
\label{DAE}

Given an image $I \in \mathbb{R}^{H \times W \times 3}$ and text prompt $t$ (Fig.~\ref{fig:mainfig}), we derive a CLIP similarity map $S \in [0,1]^{H \times W}$ as a semantic prior, while GroundingDINO simultaneously generates text-aligned object proposals. Although CLIP is pretrained with global image--text supervision, its dense features provide useful region-level semantic cues~\cite{zhou2022extract}. We use these features to measure spatial correspondence with the target text when evaluating detector proposals.

\noindent\textbf{Initial Box Selection:} Each detection box $\mathbf{b}_i$ is evaluated using the CLIP similarity response $S$ within the region. Since detections may contain multiple instances or background, we measure their internal semantic consistency using the entropy of the similarity distribution inside $\mathbf{b}_i$ (see Appendix~\ref{maths} for further details). Heterogeneous object or background content tends to produce higher entropy, whereas semantically coherent regions yield lower entropy. Let $c_i$ denote the confidence of the $i^{th}$ detection from GroundingDINO, and $\widehat{H}(\mathbf{b}_i)$ the normalized entropy of similarity values within its bounding box $\mathbf{b}_i$. Each box is assigned a composite score balancing detection confidence and the inverse entropy term:
\begin{equation}
    J_{\text{box}}(\mathbf{b}_i)
    = \alpha\, c_i
    + (1 - \alpha)\,(1 - \widehat{H}(\mathbf{b}_i)),
    \label{eq:box_score}
\end{equation} where $\alpha=0.5$ balances detection confidence and spatial purity. Higher $J_{\mathrm{box}}(\mathbf{b}_i)$ favors confident, semantically coherent regions, and the top-scoring box $\mathbf{b}^c=\arg\max_{\mathbf{b}_i}J_{\mathrm{box}}(\mathbf{b}_i)$ is selected as a coarse region of interest. Why coarse? Since even the best detector proposal may contain multiple instances, $\mathbf{b}^c$ is subsequently refined by SSES into an instance-specific region.


\noindent\textbf{Sim-guided SAM-based Exemplar Selection (SSES):}
Given $\mathbf{b}^c$, SSES refines it into an instance-specific exemplar by exploiting local peaks in $S$, which indicate regions strongly aligned with the target text. We extract up to $k$ peaks within $\mathbf{b}^c$ using an adaptive percentile threshold rather than a fixed similarity cutoff. Percentiles are computed over the full similarity map $S$, normalizing thresholds to each image’s global similarity statistics. Starting from a high percentile (e.g., 90th), the threshold is gradually relaxed in fixed steps ($\Delta p = 10$) until at most $k=16$ peaks are detected within $\mathbf{b}^{c}$ (see ablations in Sec.~\ref{experiments}). This adaptive relaxation dynamically adjusts the confidence level for peak selection based on the strength and density of responses in the coarse region. Formally, for a percentile $p$, we define the threshold as $\tau(p) = \operatorname{Perc}_p(S)$,
and retain spatial locations within $\Omega(\mathbf{b}^{c}) = \{(x, y)\,|\, x_0 \le x \le x_1,\, y_0 \le y \le y_1\}$ whose similarity exceeds this threshold:
\begin{equation}
\mathcal{Q}(p) = \{\,q = (x, y) \in \Omega(\mathbf{b}^{c}) \mid S(q) \ge \tau(p)\},
\end{equation} where $q$ denotes a local peak in $S$, and $\mathcal{Q}(p)$ represents the set of spatial locations within $\mathbf{b}^{c}$ whose similarity values surpass the global percentile threshold $\tau(p)$. This adaptive relaxation retains high-confidence peaks even when similarity contrast is weak. Each peak $\mathbf{q}_i$ is passed to SAM as a positive point prompt, producing mask $M_i=P(I;\mathbf{q}_i)$.  For each mask, we compute two complementary measures: (1) \textit{semantic strength} $\operatorname{Sim}(M_i)$, which quantifies how strongly the pixels within $M_i$ align with the text prompt $t$ via the mean percentile rank of their similarity values in $S$, i.e., $\operatorname{Sim}(M_i) = \frac{1}{|M_i|}\sum_{u \in M_i} r(u)$, where $r(u)$ is the percentile rank of $S(u)$; and (2) \textit{normalized entropy} $\widehat{H}(M_i)$, which captures how uniformly these similarity values are distributed. While $\operatorname{Sim}(M_i)$ favors text-aligned regions, $\widehat{H}(M_i)$ penalizes masks with diffuse or multi-object content. Each mask is assigned a composite score that balances the two measures:
\begin{equation}
J_{\text{mask}}(M_i) = w_{\text{sim}}\,\operatorname{Sim}(M_i) + w_{\text{ent}}\,(1 - \widehat{H}(M_i)),
\label{eq3}
\end{equation} where $w_{\text{sim}} = w_{\text{ent}} = 0.5$. The optimal mask is selected as $M_{\text{DAE}} = \arg\max_{M_i} J_{\text{mask}}(M_i)$, whose bounding box, $\mathbf{b}_{\text{DAE}}$, defines the DAE exemplar (see Appendix~\ref{SSES_vis}).




\subsection{Density-Guided Exemplar (DGE)}
\label{DGE}



 \noindent\textbf{Peak-to-Point (P2P) prompting:}
 Given $I$ and exemplar $\mathbf{b}_{\mathrm{DAE}}$, the density estimator produces a density map $D\in\mathbb{R}^{H\times W}$, with total count $C=\sum_{x,y}D(x,y)$ and local maxima indicating potential object centers (Fig.~\ref{fig:mainfig}). Rather than assuming a specific distribution for $D$, we employ the mean and standard deviation of the density map as adaptive scale statistics to identify high-confidence peaks. Specifically, an adaptive threshold is defined as $T_D = \mu_D + 2\sigma_D$, where $\mu_D$ and $\sigma_D$ denote the mean and standard deviation of $D$, respectively. Pixels satisfying $D(x,y)\geq T_D$ are retained as high-confidence candidates, and local peak detection within a fixed neighborhood ensures spatial separation, yielding $\mathcal{P}=\{\,p_i=(x_i,y_i)\mid D(p_i)\geq T_D\,\}$. To preserve semantic consistency, we retain only peaks $\mathbf{p}_i$ that fall within at least one GroundingDINO detection $\mathbf{b}_j^{\mathrm{DINO}}$. The filtered prompt set is thus
$\mathcal{P}^{\star}=\{\mathbf{p}_i\in\mathcal{P}\mid \mathbf{p}_i\in\Omega_{\mathrm{DINO}}\}$,
where $\Omega_{\mathrm{DINO}}$ is the union of all detection boxes. Each $\mathbf{p}_i\in\mathcal{P}^{\star}$ is used as a positive SAM prompt to generate mask $M_i=P(I;\mathbf{p}_i)$ and bounding box $\mathbf{b}_i$, forming the candidate exemplar pool.



\noindent\textbf{Single-instance Filtering via RoI Count:} We estimate the number of instances within each candidate $\mathbf{b}_i$ by integrating its density response as $\hat{c}(\mathbf{b}_i)=\sum_{(x,y)\in\Omega(\mathbf{b}_i)}D(x,y)$, where $\Omega(\mathbf{b}_i)$ denotes the set of pixel coordinates inside the box. Since regression noise and density diffusion may yield fractional counts above one, boxes with predicted counts satisfying $1 < \hat{c}(\mathbf{b}_i) < 2$ are retained as single-instance candidates. The resulting subset $\mathcal{B}_{\text{single}}$ represents single-instance proposals for subsequent exemplar refinement.



\begin{figure}[t]
    \centering
    \includegraphics[width=\linewidth]{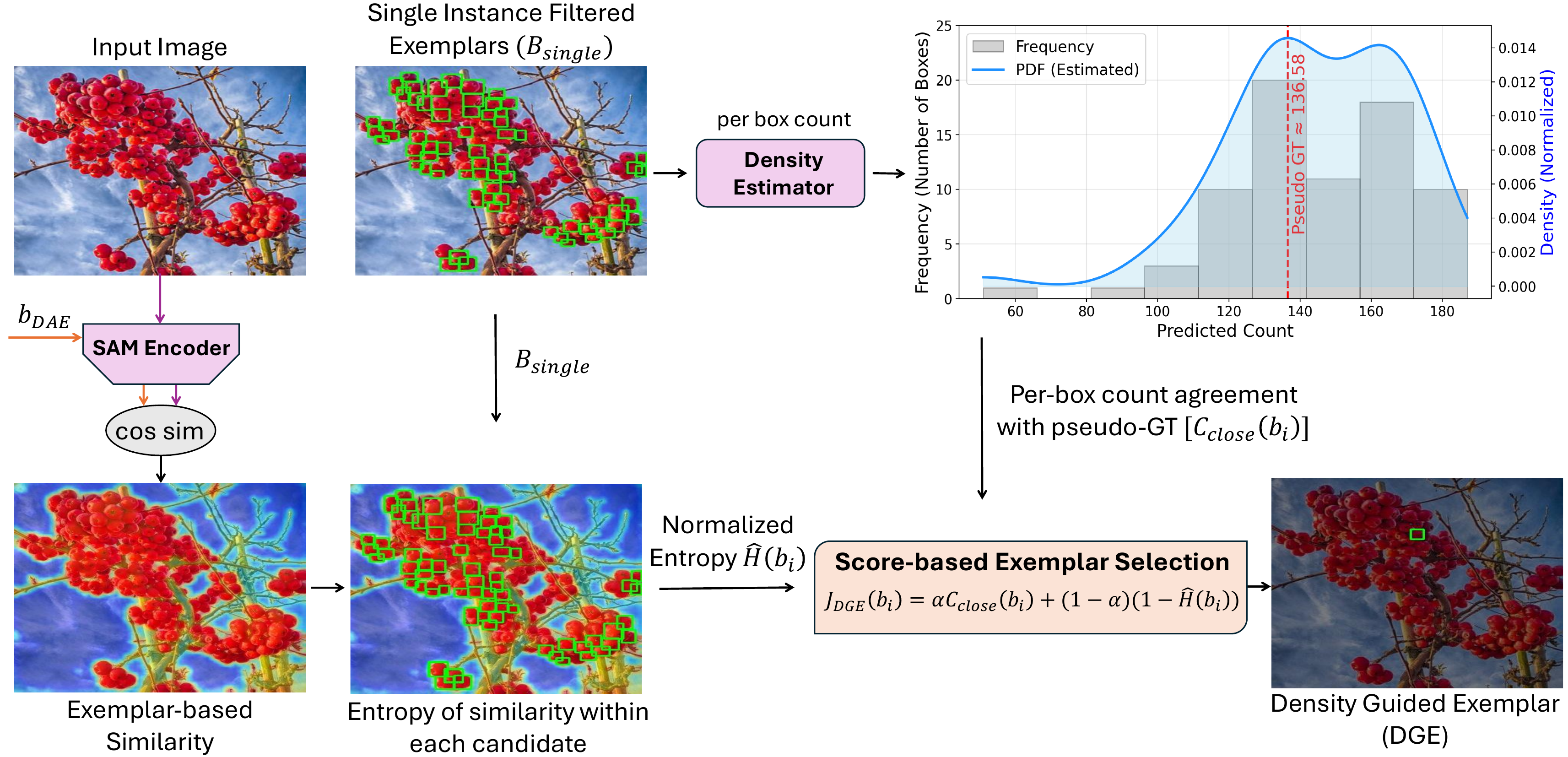}
    \caption{Overview of GGES in DGE. Per-box counts from $\mathcal{B}_{\text{single}}$ estimate a pseudo-GT, while a SAM similarity map conditioned on $\mathbf{b}_{\text{DAE}}$ captures visual correspondence. A composite score combining pseudo-GT proximity and similarity-map entropy guides density-based exemplar selection.}
    \label{fig:DGE}
\end{figure}

\noindent\textbf{Pseudo-GT Guided Exemplar Selection (GGES):}
From $\mathcal{B}_{\mathrm{single}}$, we estimate a consensus pseudo-GT count from the distribution of per-box predictions $\hat{c}_i=\hat{c}(\mathbf{b}_i)$. We model their empirical distribution using a non-parametric probability density function (PDF) (see Appendix~\ref{maths} for further details), whose dominant count region provides a robust pseudo-GT reference (Fig.~\ref{fig:DGE}). By leveraging the overall distribution instead of individual estimates, this approach mitigates outlier effects and guides exemplar selection toward the statistically most consistent estimate.

Each candidate is assigned count closeness $C_{\mathrm{close}}(\mathbf{b}_i)$ to the pseudo-GT, with higher values indicating greater statistical consistency. Since count agreement alone does not ensure semantic purity, we additionally evaluate candidate compactness using an exemplar-conditioned SAM similarity map. Unlike DAE, which relies on text-conditioned CLIP similarity due to the absence of a concrete exemplar, DGE leverages $\mathbf{b}_{\mathrm{DAE}}$ to obtain an instance-specific similarity map. We encode $\mathbf{b}_{\mathrm{DAE}}$ with SAM and compute cosine similarity against feature embeddings of the entire image, highlighting regions that share strong visual and structural correspondence with the exemplar (Fig.~\ref{fig:DGE}). The normalized entropy of this similarity map within each box $\mathbf{b}_i$, denoted as $\widehat{H}(\mathbf{b}_i)$, quantifies how spatially concentrated the exemplar’s feature activation is. Low entropy indicates compact, object-specific responses, while high entropy suggests background interference or multiple objects. Thus, $\widehat{H}(\mathbf{b}_i)$ complements $C_{\text{close}}(\mathbf{b}_i)$ by ensuring the exemplar is both statistically consistent and semantically pure. Each candidate is scored as:
\begin{equation}
J_{\text{DGE}}(\mathbf{b}_i) =
\alpha\, C_{\text{close}}(\mathbf{b}_i)
+ (1-\alpha)\,(1-\widehat{H}(\mathbf{b}_i)),
\label{eq6}
\end{equation} where $\alpha=0.5$ balances count consistency and semantic compactness. The highest-scoring candidate,
$\mathbf{b}_{\mathrm{DGE}}=\arg\max_{\mathbf{b}_i\in\mathcal{B}_{\mathrm{single}}}J_{\mathrm{DGE}}(\mathbf{b}_i)$,
defines DGE exemplar. 



\begin{figure}[t]
    \centering
    \includegraphics[width=\linewidth]{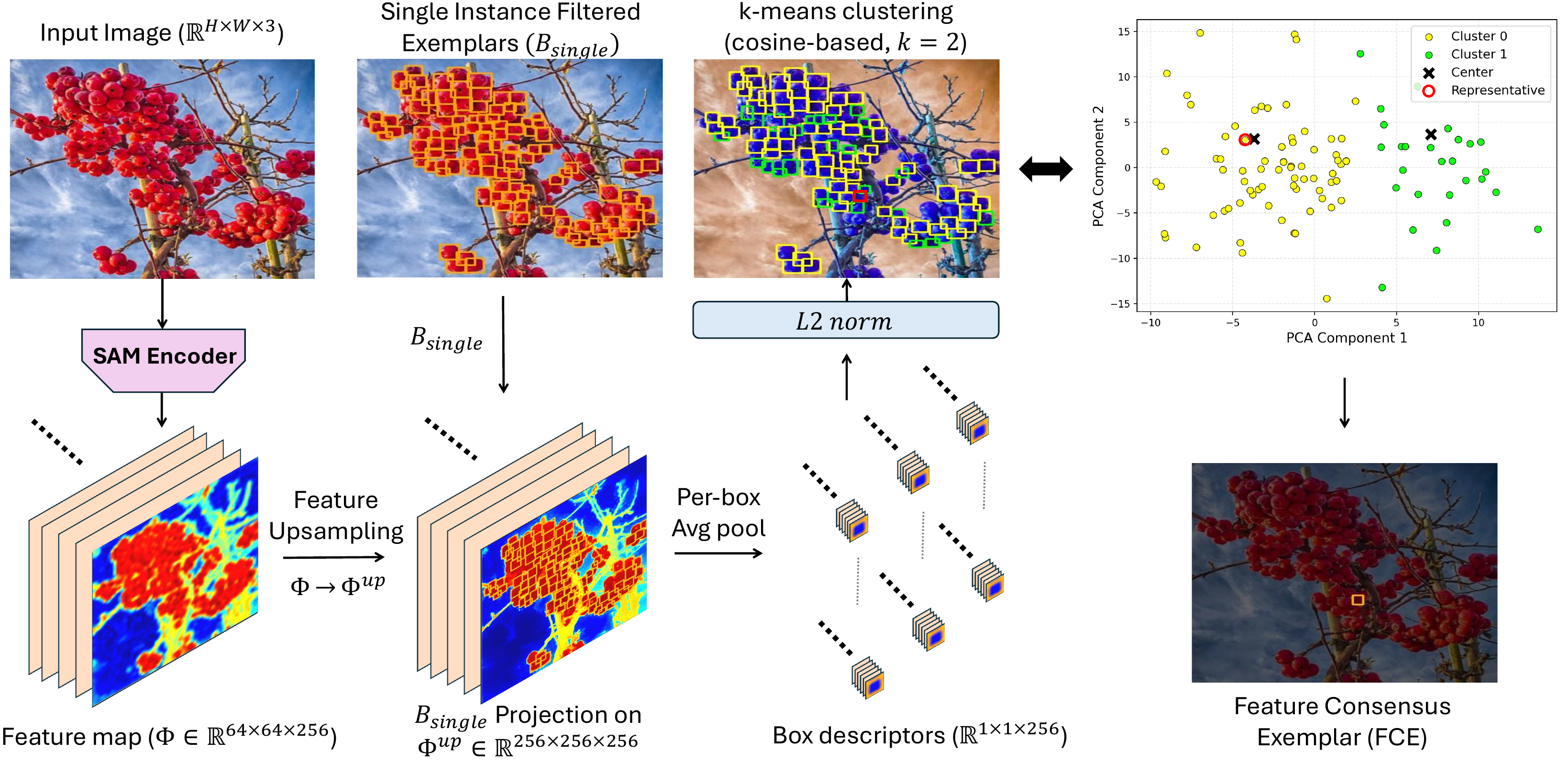}
    \caption{Overview of FRES in the FCE stage. Single-instance boxes $\mathcal{B}_{\text{single}}$ are projected onto the upsampled SAM feature map $\Phi^{\text{up}}$ to obtain $\ell_2$-normalized regional embeddings. These embeddings are grouped via binary clustering, and the box closest to the majority-cluster centroid is selected as the representative exemplar.}
    \label{fig:FCE}
\end{figure}

\subsection{Feature-Consensus Exemplar (FCE)}
\label{FCE}

DGE enforces statistical consistency but does not explicitly capture visual coherence among candidate instances. Using $\mathbf{b}_{\mathrm{DGE}}$ as the anchor, FCE repeats density estimation, P2P prompting, and single-instance filtering to obtain $\mathcal{B}_{\mathrm{single}}$, then replaces GGES with FRES (discussed next) to identify the dominant visual appearance.





\noindent\textbf{Feature-based Representative Exemplar Selection (FRES):} Given $\mathcal{B}_{\mathrm{single}}$, FRES selects an exemplar representing the dominant object appearance (Fig.~\ref{fig:FCE}). We encode $I$ with SAM to obtain $\Phi\in\mathbb{R}^{256\times64\times64}$ and upsample it to $\Phi^{\mathrm{up}}\in\mathbb{R}^{256\times256\times256}$ using window-attention-based upsampling~\cite{wimmer2025anyup}, reducing spatial detail loss for small candidates. Each candidate box $\mathbf{b}_i$ is then projected onto $\Phi^{\text{up}}$, and the corresponding region $\Phi^{\text{up}}(\Omega_f(\mathbf{b}_i))$ defines its local feature subset. Finally, each box descriptor $\mathbf{v}_i$ is computed via spatial average pooling: 
\begin{equation}
\mathbf{v}_i =
\frac{1}{|\Omega_f(\mathbf{b}_i)|}\sum{(u,v)\in \Omega_f(\mathbf{b}_i)} \Phi^{\text{up}}(:,v,u),
\end{equation} The vector $\mathbf{v}_i$ thus represents the semantic appearance of region $\mathbf{b}_i$ in the SAM feature space. Since the magnitude of pooled features may vary with object size, texture, or illumination, direct comparison between descriptors can be biased toward high-energy regions. To mitigate this effect, we apply $\ell_2$-normalization so that all descriptors lie on a unit hypersphere, ensuring that their pairwise dot products correspond to cosine similarity.


While each descriptor $\mathbf{v}_i$ captures the visual characteristics of a single-instance region, these descriptors may still vary due to pose, scale, illumination, or minor background interference.  We therefore perform binary clustering (majority and a minority cluster) in feature space to separate the dominant appearance from peripheral or noisy candidates. Let $\mathcal{C}_{\mathrm{maj}}$ denote the majority cluster and $\boldsymbol{\mu}_{\mathrm{maj}}$ its centroid. The representative exemplar is selected as
$\mathbf{b}_{\mathrm{FCE}}=\arg\max_{\mathbf{b}_i\in\mathcal{C}_{\mathrm{maj}}}s(\mathbf{v}_i,\boldsymbol{\mu}_{\mathrm{maj}})$,
where $s(\cdot,\cdot)$ denotes cosine similarity. The selected $\mathbf{b}_{\mathrm{FCE}}$ captures the dominant feature-level consensus, complementing the count-driven DGE exemplar. Finally, $\{\mathbf{b}_{\mathrm{DAE}},\mathbf{b}_{\mathrm{DGE}},\mathbf{b}_{\mathrm{FCE}}\}$ forms the complementary exemplar set forwarded to density estimator.

\section{Experiments and Ablations}
\label{experiments}

\begin{table}[t]
\centering
\setlength{\tabcolsep}{1.8pt}
\fontsize{7.6pt}{7.6pt}\selectfont
\caption{Comparison of CountZES with SOTA ZOC methods on FSC-147 dataset. Best results are in bold and second-best are underlined. $^\dagger$ indicates ZOC methods based on exemplar discovery. * indicates that CountingDINO is not zero-shot (box prompts).}
\begin{tabular}{llccc}
\toprule
\rowcolor{lightgray!30} 
\multirow{2}{*}{Method} 
& \multirow{2}{*}{Venue} 
& \multicolumn{1}{c}{Inference} 
& \multirow{2}{*}{MAE} 
& \multirow{2}{*}{RMSE} \\
&  & \multicolumn{1}{c}{only} &  &  \\
\midrule
\rowcolor{lightgreen!40}
ZSOC$^\dagger$~\cite{xu2023zero} & CVPR'23 & \xmark & 22.09 & 115.17 \\
CLIP-Count~\cite{jiang2023clip} & ACM'23 & \xmark & 17.78 & 106.62 \\
CounTX~\cite{amini2023open} & BMVC'23 & \xmark & 15.88 & 106.29 \\
VLCounter~\cite{kang2024vlcounter} & AAAI'24 & \xmark & 17.05 & 106.16 \\
PseCo~\cite{huang2024point} & CVPR'24 & \xmark & 16.58 & 129.77 \\
DAVE~\cite{pelhan2024dave} & ICCV'24 & \xmark & 14.90 & 103.42 \\
\rowcolor{lightgreen!40}
VA-Count$^\dagger$~\cite{zhu2024zero} & ECCV'24 & \xmark & 17.88 & 129.31 \\
CountGD~\cite{amini2024countgd} & NeurIPS'24 & \xmark & 14.76 & 120.42 \\
GeCo~\cite{pelhan2024novel} & NeurIPS'24 & \xmark & 13.30 & 108.72 \\
T2ICount~\cite{qian2025t2icount} & CVPR'25 & \xmark & \underline{11.76} & 97.86 \\
CountSE~\cite{liu2025countse}& ICCV'25 & \xmark & \textbf{7.84} & \textbf{82.99} \\
CountGD++~\cite{amini2026countgd++}& CVPR'26 & \xmark & 16.55 & 129.76 \\
\rowcolor{goldenbrown!30}
CountZES$^\dagger$ \textit{(in-domain)} & (Ours) & \xmark & 15.77 & \underline{91.40} \\

\midrule
CountingDINO*~\cite{pacini2025countingdino} & WACV'26 & \cmark & 20.93 & 71.37 \\
\midrule
SAM~\cite{kirillov2023segment} & ICCV'23 & \cmark & 42.48 & 137.50 \\
Count Anything~\cite{ma2023can} & arXiv'23 & \cmark & 27.97 & 131.24 \\
G-DINO~\cite{liu2024grounding} & ECCV'24 & \cmark & 59.23 & 159.28 \\
TFOC~\cite{shi2024training} & WACV'24 & \cmark & 24.79 & 137.15 \\
TFCAC~\cite{lin2025simple} & WACV'25 & \cmark & \underline{23.59} & \underline{113.60} \\
OmniCount~\cite{mondal2025omnicount} & AAAI'25 & \cmark & 21.46 & 133.28 \\
SAM3~\cite{carion2025sam} & ICLR'26 & \cmark & 22.89 & 144.80 \\
\rowcolor{goldenbrown!30}
CountZES$^\dagger$ \textit{(cross-domain)} & (Ours) & \cmark & \textbf{21.09} & \textbf{110.14} \\
\bottomrule
\end{tabular}
\label{tab:fsc147_eval}
\vspace{-3pt}
\end{table}

\noindent\textbf{Datasets and Evaluation Metrics.}
We evaluate CountZES on FSC-147~\cite{ranjan2021learning} (6,135 images, 147 classes), CARPK~\cite{hsieh2017drone} (1,448 aerial images, 89,777 cars), and the dense PerSense-D~\cite{siddiqui2025persense} benchmark (717 images, 28 categories, 53 instances/image on average). Cross-domain medical evaluation uses VGG~\cite{lempitsky2010learning} (200 fluorescence microscopy images) and MBM~\cite{paul2017count} (44 microscopy images, $126\pm33$ nuclei/image). Following prior ZOC works, we report mean absolute error (MAE) and root mean square error (RMSE).

\noindent\textbf{Implementation Details.} CountZES employs CLIP~\cite{radford2021learning} for text--image alignment, GroundingDINO~\cite{liu2024grounding} for initial proposals with a detection threshold of 0.15, and SAM~\cite{kirillov2023segment} for point-prompted mask generation. To demonstrate the modular nature, we employ CounTR~\cite{liu2022countr} and DSALVANet~\cite{he2024few} as pretrained exemplar-based density estimators (DEs); alternative segmentation backbones are also examined in the ablations. We distinguish two evaluation settings according to the pretraining source of the DE: \textit{in-domain} uses a frozen DE pretrained on the target dataset, whereas \textit{cross-domain} uses a frozen DE pretrained on a different dataset. CountZES performs no ZOC-specific optimization, fine-tuning, or test-time adaptation in either setting. For conservative reporting, Table~\ref{tab:fsc147_eval} marks only the cross-domain setting as inference-only because the in-domain DE is pretrained on target dataset.

\noindent\textbf{Quantitative Results on FSC-147.} Table~\ref{tab:fsc147_eval} evaluates CountZES under in-domain and cross-domain DE settings. With DSALVANet pretrained on FSC-147 but frozen during inference, CountZES reduces MAE/RMSE by 28.6\%/20.6\% over ZSOC and 11.8\%/29.3\% over VA-Count, highlighting its effectiveness over existing visual-exemplar discovery approaches. In the cross-domain setting, using a DE pretrained on CARPK, CountZES achieves the best MAE and RMSE among the compared inference-only zero-shot methods, demonstrating effective transfer without target-domain counter pretraining or ZOC-specific adaptation.


\begin{table}[t]
\centering
\setlength{\tabcolsep}{2pt}
\fontsize{7.6pt}{7.6pt}\selectfont
\caption{CountZES compared to SOTA on CARPK dataset.}
\begin{tabular}{lccc}
\toprule
\rowcolor{lightgray!30}
Method & Inference-only & MAE & RMSE \\
\midrule
CLIP-Count~\cite{jiang2023clip} & \xmark & 11.96 & 16.61 \\
CounTX~\cite{amini2023open} & \xmark & 8.13 & 10.87 \\
G-DINO~\cite{liu2024grounding} & \xmark & 29.72 & 31.60 \\
VLCounter~\cite{kang2024vlcounter} & \xmark & 6.46 & 8.68 \\
\rowcolor{lightgreen!40}
VA-Count$^\dagger$~\cite{zhu2024zero} & \xmark & 10.63 & 13.20 \\ 
GeCo~\cite{pelhan2024novel} & \xmark & 10.34 & 14.73 \\
CountGD~\cite{amini2024countgd} & \xmark & 3.83 & 5.41 \\
T2ICount~\cite{qian2025t2icount} & \xmark & 8.61 & 13.47 \\
CountSE~\cite{liu2025countse} & \xmark & \underline{2.79} & \underline{4.20} \\
CountGD++~\cite{amini2026countgd++} & \xmark & \textbf{2.48} & \textbf{3.74} \\
\midrule
TFOC~\cite{shi2024training} & \cmark & 14.35 & 17.22 \\
OmniCount~\cite{mondal2025omnicount}  & \cmark & \underline{13.41} & \underline{16.85} \\
CountingDINO~\cite{pacini2025countingdino} & \cmark & 21.26 & 28.20 \\
\rowcolor{goldenbrown!30}
CountZES$^\dagger$ (Ours) & \cmark & \textbf{7.24} & \textbf{9.22} \\
\bottomrule
\end{tabular}
\label{tab:carpk_eval}
\end{table}

\noindent\textbf{Quantitative Results on CARPK.} Using DSALVANet pretrained on FSC-147, CountZES achieves the lowest MAE and RMSE among inference-only approaches on CARPK (Table~\ref{tab:carpk_eval}), while remaining competitive with several ZOC-optimized methods. Compared with the visual-exemplar-based VA-Count, it reduces MAE/RMSE by 31.9\%/30.2\%. These results demonstrate strong cross-domain transfer of the proposed multi-stage exemplar discovery and selection framework without ZOC-specific optimization.

\begin{table}[t]
\centering
\caption{CountZES compared to SOTA on PerSense-D dataset.}
\fontsize{7pt}{7pt}\selectfont
\setlength{\tabcolsep}{1.3pt}
\begin{tabular}{l c c c c c c | c c}
\toprule
\rowcolor{lightgray!30}
\multirow{3}{*}{} 
& \multicolumn{6}{c}{Density-based} 
& \multicolumn{2}{c}{Class-wise} \\
\cmidrule(lr){2-7}\cmidrule(lr){8-9}
\rowcolor{lightgray!30}
\multirow{-1}{*}{Method (Inf-only)} 
& \multicolumn{2}{c}{Low} & \multicolumn{2}{c}{Med} & \multicolumn{2}{c}{High} & \multicolumn{2}{c}{Overall} \\
\cmidrule(lr){2-3}\cmidrule(lr){4-5}\cmidrule(lr){6-7}\cmidrule(lr){8-9}
\rowcolor{lightgray!30}
& MAE & RMSE & MAE & RMSE & MAE & RMSE & MAE & RMSE  \\
\midrule
GeCo (\xmark) & \underline{11.63} & \underline{17.84} & \underline{10.30} & \underline{20.61} & \textbf{12.38} & \textbf{24.13} & \underline{13.12} & \underline{28.36} \\
\addlinespace[0.3em]
T2ICount (\xmark) & 30.04 & 122.91 & 25.16 & 76.95 & 20.62 & 52.23 & 24.95 & 88.21 \\
\addlinespace[0.3em]

\midrule
TFOC (\cmark) & 7.20 & \textbf{10.36} & 16.07 & 21.63& 36.88 & 63.82 & 17.79 & 35.74 \\

\rowcolor{goldenbrown!30}
CountZES (\cmark) & \textbf{6.86} & 19.22 & \textbf{10.26} & \textbf{14.44} & \underline{20.36} & \underline{43.65} & \textbf{12.29} & \textbf{28.32} \\
\bottomrule
\end{tabular}
\label{tab:persense_density_classwise}
\vspace{-3pt}
\end{table}

\begin{table}[t]
\centering
\caption{Comparison of CountZES with SOTA ZOC methods on medical cell counting datasets VGG and MBM.}
\setlength{\tabcolsep}{5pt}
\fontsize{7pt}{7pt}\selectfont
\begin{tabular}{l c c c c}
\rowcolor{lightgray!30}
\toprule
\multirow{1}{*}{Method (Inf-only)} & \multicolumn{2}{c}{VGG} & \multicolumn{2}{c}{MBM} \\
\cmidrule(lr){2-3}\cmidrule(lr){4-5}\rowcolor{lightgray!30}
 & MAE & RMSE & MAE & RMSE \\
\midrule
GeCo (\xmark) & 67.32 & 77.53 & 92.59 & 104.52 \\
T2ICount (\xmark) & 151.35 & 162.81 & \underline{30.01} & \underline{44.44} \\
CountSE (\xmark) & \underline{64.75} & \underline{74.51} & 61.90 & 69.97 \\
CountGD++ (\xmark) & 173.70 & 183.74 & 121.50 & 125.76 \\

\midrule
TFOC (\cmark) & 144.91 & 158.52 & 64.25 & 75.67  \\
\rowcolor{goldenbrown!30}
CountZES (\cmark) & \textbf{45.55} & \textbf{74.41} & \textbf{22.16} & \textbf{40.06} \\
\bottomrule
\end{tabular}
\label{tab:vgg_mbm_comparison}
\vspace{-3pt}
\end{table}

\begin{table}[t]
\centering
\caption{Ablation on FSC-147.
\textbf{Left:} Effect of stage interaction and ordering.
\emph{Parallel} independently conditions both DGE and FCE on DAE.
\emph{Seq.~1} follows DAE$\rightarrow$FCE$\rightarrow$DGE, while
\emph{Seq.~2} follows DAE$\rightarrow$DGE$\rightarrow$FCE, propagating the selected exemplar sequentially between stages.
\textbf{Right:} Stage-wise contribution.}
\label{tab:medical_ablation_combined}


\begin{minipage}[t]{0.61\linewidth}
\centering
\setlength{\tabcolsep}{1.1pt}
\fontsize{6.5pt}{7pt}\selectfont

\begin{tabular}{lcc|cc|cc}
\toprule
\rowcolor{lightgray!30}
\multirow{2}{*}{Setting}
& \multicolumn{2}{c|}{Parallel}
& \multicolumn{2}{c|}{Seq.~1}
& \multicolumn{2}{c}{Seq.~2} \\
\cmidrule(lr){2-3}
\cmidrule(lr){4-5}
\cmidrule(lr){6-7}
\rowcolor{lightgray!30}
& MAE & RMSE
& MAE & RMSE
& MAE & RMSE \\
\midrule

In-domain
& 17.12 & 92.57
& 16.58 & 91.90
& \cellcolor{goldenbrown!30}15.77
& \cellcolor{goldenbrown!30}91.40 \\

Cross-domain
& 22.31 & 112.01
& 21.87 & 111.23
& \cellcolor{goldenbrown!30}21.09
& \cellcolor{goldenbrown!30}110.14 \\

\bottomrule
\end{tabular}

\end{minipage}
\hfill
%
\begin{minipage}[t]{0.36\linewidth}
\centering
\setlength{\tabcolsep}{1.1pt}
\fontsize{6.5pt}{7pt}\selectfont

\begin{tabular}{ccccc}
\toprule
\rowcolor{lightgray!30}
DAE & DGE & FCE & MAE & RMSE \\
\midrule
\on & \off & \off & 25.13 & 97.40 \\
\on & \on  & \off & 20.75 & 92.36 \\
\rowcolor{goldenbrown!30}
\on & \on & \on & 15.77 & 91.40 \\
\bottomrule
\end{tabular}

\end{minipage}

\vspace{-4pt}
\end{table}

\begin{table*}[t]
\centering
\caption{Ablation studies of CountZES.
\textbf{(a)} Advantage of cross-stage collaboration over single-stage exemplar selection.
\textbf{(b)} Robustness to poor detector initialization.
\textbf{(c)} Effect of the number of peaks $k$ in SSES.
\textbf{(d)} Effect of density-guided candidate discovery in DGE.
\textbf{(e)} Flexible design with different segmentation backbones.}
\label{tab:combined_ablations}
\vspace{-4pt}

\begin{minipage}[t]{0.16\textwidth}
\centering
\phantomsubcaption
\label{tab:cross_stage}
\setlength{\tabcolsep}{1.5pt}
\fontsize{6pt}{6.5pt}\selectfont

\textbf{(a) Cross-stage collaboration}\\[2pt]

\resizebox{\linewidth}{!}{%
\begin{tabular}{lcc}
\toprule
\rowcolor{lightgray!30}
Configuration & MAE & RMSE \\
\midrule
DAE only & 21.73 & 118.03 \\
DGE only & 17.38 & 108.56 \\
FCE only & 19.42 & 116.53 \\
\rowcolor{goldenbrown!30}
CountZES & 15.77 & 91.40 \\
\bottomrule
\end{tabular}%
}
\end{minipage}%
\hfill%
\begin{minipage}[t]{0.21\textwidth}
\centering
\phantomsubcaption
\label{tab:detector_proposals}
\setlength{\tabcolsep}{1.5pt}
\fontsize{6.5pt}{6.3pt}\selectfont

\textbf{(b) Detector proposals}\\[2pt]

\resizebox{\linewidth}{!}{%
\begin{tabular}{lcc|cc}
\toprule
\rowcolor{lightgray!30}
\multirow{2}{*}{Variant}
& \multicolumn{2}{c|}{FSC-147}
& \multicolumn{2}{c}{CARPK} \\
\cmidrule(lr){2-3}\cmidrule(lr){4-5}
\rowcolor{lightgray!30}
& MAE & RMSE & MAE & RMSE \\
\midrule
\rowcolor{goldenbrown!30}
Original   & 21.09 & 110.14 & 7.24 & 9.22 \\
Worst-case & 23.05 & 113.57 & 8.82 & 10.91 \\
\bottomrule
\end{tabular}%
}
\end{minipage}%
\hfill%
\begin{minipage}[t]{0.18\textwidth}
\centering
\phantomsubcaption
\label{tab:sses_peaks}
\setlength{\tabcolsep}{1.5pt}
\fontsize{6pt}{6.5pt}\selectfont

\textbf{(c) Peaks in SSES}\\[2pt]

\resizebox{\linewidth}{!}{%
\begin{tabular}{lcc|cc}
\toprule
\rowcolor{lightgray!30}
\multirow{2}{*}{$k$}
& \multicolumn{2}{c|}{FSC-147}
& \multicolumn{2}{c}{CARPK} \\
\cmidrule(lr){2-3}\cmidrule(lr){4-5}
\rowcolor{lightgray!30}
& MAE & RMSE & MAE & RMSE \\
\midrule
4  & 16.41 & 91.47 & 7.73 & 9.41 \\
8  & 16.22 & 91.65 & 7.37 & 9.40 \\
\rowcolor{goldenbrown!30}
16 & 15.77 & 91.40 & 7.24 & 9.22 \\
\bottomrule
\end{tabular}%
}
\end{minipage}%
\hfill%
\begin{minipage}[t]{0.16\textwidth}
\centering
\phantomsubcaption
\label{tab:density_candidate_ablation}
\setlength{\tabcolsep}{1.2pt}
\fontsize{6pt}{6.5pt}\selectfont

\textbf{(d) DGE candidate discovery}\\[2pt]

\resizebox{\linewidth}{!}{%
\begin{tabular}{lcc}
\toprule
\rowcolor{lightgray!30}
Candidate pool & MAE & RMSE \\
\midrule
Detector only
& 18.67 & 116.30 \\
\rowcolor{goldenbrown!30}
Density-guided
& 15.77 & 91.40 \\
\bottomrule
\end{tabular}%
}
\end{minipage}%
\hfill%
\begin{minipage}[t]{0.20\textwidth}
\centering
\phantomsubcaption
\label{tab:segmentation_backbones}
\setlength{\tabcolsep}{1.3pt}
\fontsize{6pt}{6.5pt}\selectfont

\textbf{(e) Segmentation backbones}\\[2pt]

\resizebox{\linewidth}{!}{%
\begin{tabular}{lcc|cc}
\toprule
\rowcolor{lightgray!30}
\multirow{2}{*}{Variant}
& \multicolumn{2}{c|}{FSC-147}
& \multicolumn{2}{c}{CARPK} \\
\cmidrule(lr){2-3}\cmidrule(lr){4-5}
\rowcolor{lightgray!30}
& MAE & RMSE & MAE & RMSE \\
\midrule
SAM
& 21.09 & 110.14 & 7.24 & 9.22 \\
SAM-HQ
& 20.94 & 110.03 & 7.13 & 9.07 \\
\rowcolor{goldenbrown!30}
SAPNet
& 19.16 & 106.76 & 6.31 & 8.28 \\
\bottomrule
\end{tabular}%
}
\end{minipage}

\vspace{-1pt}
\end{table*}

\begin{figure*}[t]
    \centering
    \includegraphics[width=0.85\linewidth]{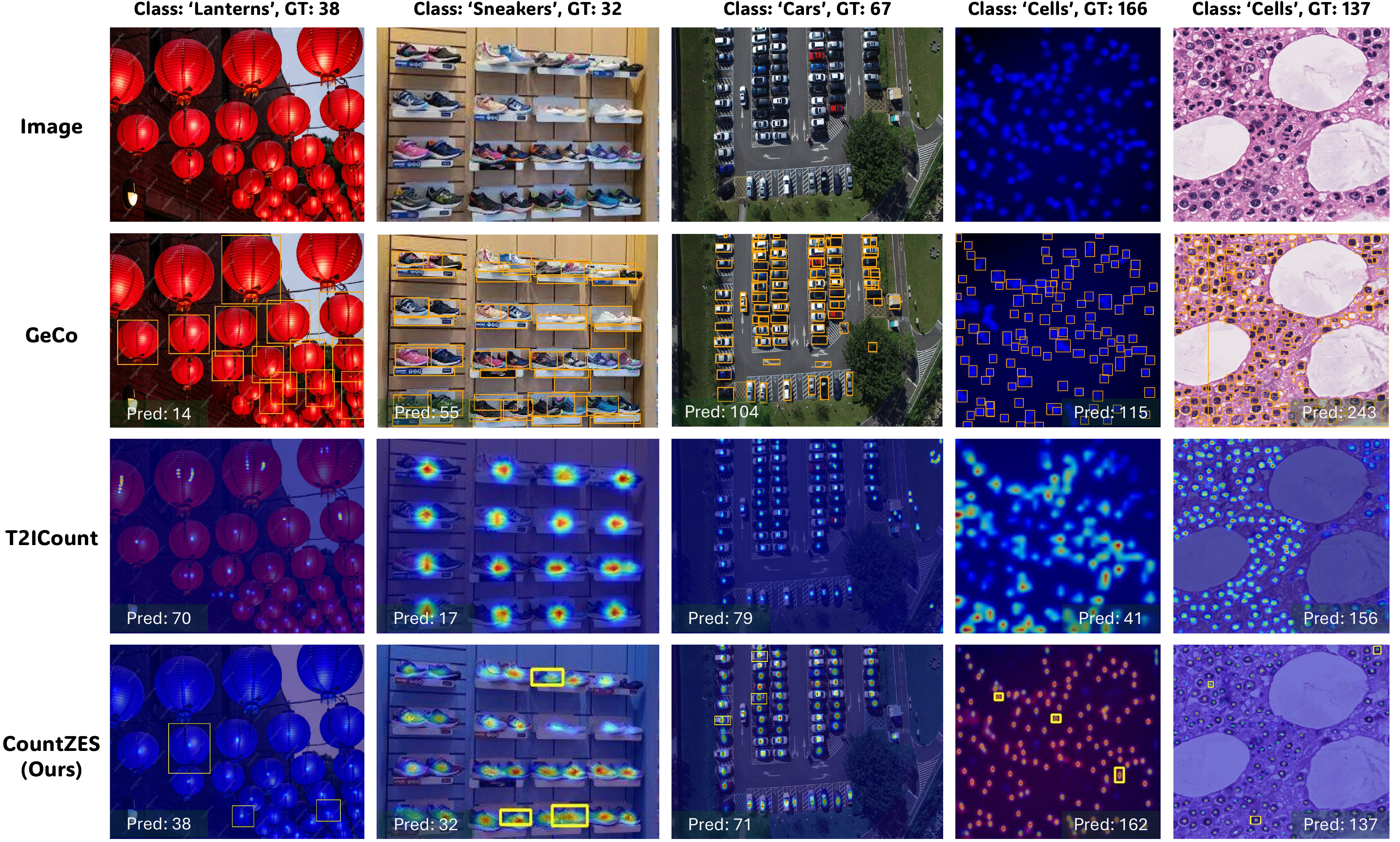}
    \caption{Qualitative comparison of CountZES with T2ICount~\cite{qian2025t2icount} and GeCo~\cite{pelhan2024novel} across diverse scenarios, including natural (FSC-147 and PerSense-D), aerial (CARPK), and medical cell-counting (VGG and MBM) images.}
    \label{fig:qualitative}
    \vspace{-3pt}
\end{figure*}

\noindent\textbf{Quantitative Results on PerSense-D.} Following the same cross-dataset protocol, we evaluate CountZES using CounTR pretrained on FSC-147 and report overall performance and the \textit{low}, \textit{medium}, and \textit{high} density splits defined in~\cite{siddiqui2025persense} (Table~\ref{tab:persense_density_classwise}). CountZES consistently achieves competitive performance across varying crowding conditions, with particularly strong gains in medium and high-density scenes. Notably, despite 16 of 28 PerSense-D classes ($\approx$57\%) overlapping with FSC-147, T2ICount exhibits substantial degradation from its strong in-domain FSC-147 performance, indicating that category overlap alone does not ensure robust transfer under shifts in appearance, scale, and density. We attribute this difference in part to how textual guidance is incorporated. T2ICount is explicitly optimized for text-guided counting, coupling textual representations with numerical estimation, whereas CountZES uses text only to initialize query-specific visual exemplar discovery while counting remains visual and exemplar-driven. This separation allows the target representation to be rediscovered from each query image, providing greater flexibility under distribution shift without ZOC-specific optimization.

\noindent\textbf{Quantitative Results on VGG and MBM.} We further evaluate cross-domain transfer on the VGG and MBM medical datasets. CountZES achieves the best MAE and RMSE among the compared methods on both datasets, including ZOC-trained approaches (Table~\ref{tab:vgg_mbm_comparison}), with particularly strong gains on MBM despite its high appearance variability and crowded nuclei. Together with CARPK and PerSense-D, these results demonstrate that query-specific exemplar discovery effectively connects textual specification with a frozen exemplar-based counter across substantial domain shifts. Fig.~\ref{fig:qualitative} and Supp A.5 present qualitative examples.

\begin{figure*}[h]
    \centering
    \includegraphics[width=0.90\linewidth]{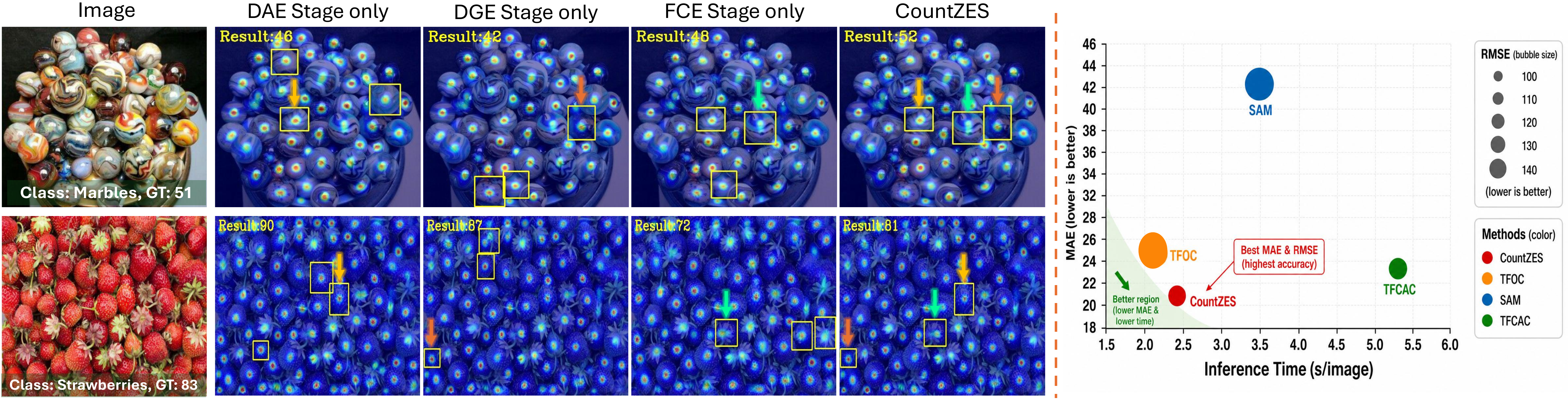}
    \caption{ \textbf{Left:} For each stage, the top three exemplars are shown along with their predicted counts. One exemplar from each stage (arrows) provides complementary cues, and their combination in CountZES improves overall counting performance (last column). \textbf{Right:} Performance–runtime tradeoff among inference-only zero-shot counting methods. (best viewed in zoom).}
    \label{fig:stagewisecomp}
    \vspace{-7pt}
\end{figure*}

\noindent\textbf{Ablation Study. } \textbf{\textit{(a) Role of stage interaction and ordering.}} Table~\ref{tab:medical_ablation_combined} (left) shows that sequential conditioning outperforms the \emph{Parallel} setting, where DGE and FCE are independently conditioned on DAE. Moreover, DAE$\rightarrow$DGE$\rightarrow$FCE consistently outperforms the reverse ordering. Placing DGE before FCE allows feature consensus to operate on a richer density-guided candidate pool generated from the statistically consistent DGE exemplar, rather than directly from the detector-constrained DAE anchor. This supports the proposed progression from semantic grounding to statistical consistency and feature-level consensus. \textbf{\textit{(b) Stage-wise contribution.}} The progressive ablation in Table~\ref{tab:medical_ablation_combined} (right) shows consistent accuracy improvements as DGE and FCE are successively added to DAE, validating their complementary contributions. \textbf{\textit{(c) Cross-stage exemplar diversity.}}
With the exemplar budget fixed to three, selecting one exemplar from each stage consistently outperforms selecting the top three from any single stage (Table~\ref{tab:combined_ablations}\subref{tab:cross_stage}). Fig.~\ref{fig:stagewisecomp} further shows that the stages capture different valid appearances and induce distinct density responses. This confirms that the gain arises from complementary semantic, statistical, and feature-level perspectives rather than simply using more exemplars. \textbf{\textit{(d) Robustness to poor detector initialization.}} To assess CountZES's dependence on the initial detector proposals, we consider a worst-case setting that bypasses DAE initial box selection and forwards the entire image as the coarse region to SSES.  Despite this worst-case initialization, CountZES incurs only a moderate performance drop on FSC-147 and CARPK (Table~\ref{tab:combined_ablations}\subref{tab:detector_proposals}), indicating that subsequent exemplar refinement and density-guided discovery can still recover from poor detector initialization. \textbf{\textit{(e) Sensitivity to design choices.}} Increasing the maximum number of SSES peaks $k$ improves performance across FSC-147 and CARPK, with $k=16$ performing best and used throughout (Table~\ref{tab:combined_ablations}\subref{tab:sses_peaks}). Appendix~\ref{design_choices_appendix} also analyzes the composite-score weight $\alpha$ ($0.5$ fixed across datasets), binary clustering in FRES, and alternative feature-descriptor definitions. \textbf{\textit{(f) Effect of density-guided candidate discovery.}}
We restrict DGE to the initial GroundingDINO proposals while keeping RoI-based filtering, GGES, FCE, and the final exemplar budget unchanged. This restriction degrades counting performance (Table~\ref{tab:combined_ablations}\subref{tab:density_candidate_ablation}), demonstrating that density-guided discovery provides a more effective candidate space for exemplar selection rather than relying solely on detector proposals. \textbf{\textit{(g) Modularity of segmentation backbone.}} Replacing SAM with SAM-HQ consistently improves performance, while SAPNet further reduces error on both FSC-147 and CARPK (Table~\ref{tab:combined_ablations}\subref{tab:segmentation_backbones}). Thus, CountZES supports stronger pretrained segmentation backbones without modifying the pipeline. 

\noindent\textbf{Runtime Efficiency.} CountZES is evaluated on FSC-147 using an NVIDIA RTX A5000 GPU. Fig.~\ref{fig:stagewisecomp} (right) illustrates the tradeoff between inference time and accuracy. Despite its progressive multi-stage design, CountZES maintains competitive inference efficiency. It requires 2.3s/image, close to TFOC (2.1s) and faster than SAM (3.4s) and TFCAC (5.26s), while achieving substantially lower counting error.

\noindent\textbf{Limitations.} CountZES relies on pretrained components and is therefore influenced by their performance. Under domain shift, DE mismatch can produce noisy density responses that propagate through progressive exemplar discovery. We mitigate this through detector-gated peaks, RoI-based single-instance filtering, consensus-based pseudo-GT estimation, and complementary count-consistency and feature-similarity criteria. Nevertheless, severe density errors may propagate when most candidates are unreliable, while SAM may produce partial or background-contaminated regions due to limited semantic awareness (see Appendix~\ref{qualitative_failure} for failure cases). However, the modular design allows stronger pretrained components to be incorporated, leading to performance improvements as demonstrated in Table~\ref{tab:combined_ablations}\subref{tab:segmentation_backbones}.

\vspace{-1pt}
\section{Conclusion}

\label{conclusion}

We presented CountZES, an inference-only framework for ZOC that addresses both visual exemplar discovery and selection. Rather than restricting exemplar selection to random patches or detector-constrained proposals, CountZES leverages density responses to expand the candidate pool. Its three-stage design then combines semantic grounding, density-guided statistical consistency, and feature-level consensus to select a set of rich exemplars. CountZES requires no ZOC-specific optimization and adopts a modular design that supports different pretrained components. Extensive evaluations across natural, aerial, and medical domains demonstrate strong cross-domain generalization and competitive performance among inference-only ZOC methods.

{
    \small
    \bibliographystyle{ieeenat_fullname}
    \bibliography{main}
}

\clearpage

\appendix

\begin{figure*}[t]
    \centering
    \includegraphics[width=1\linewidth]{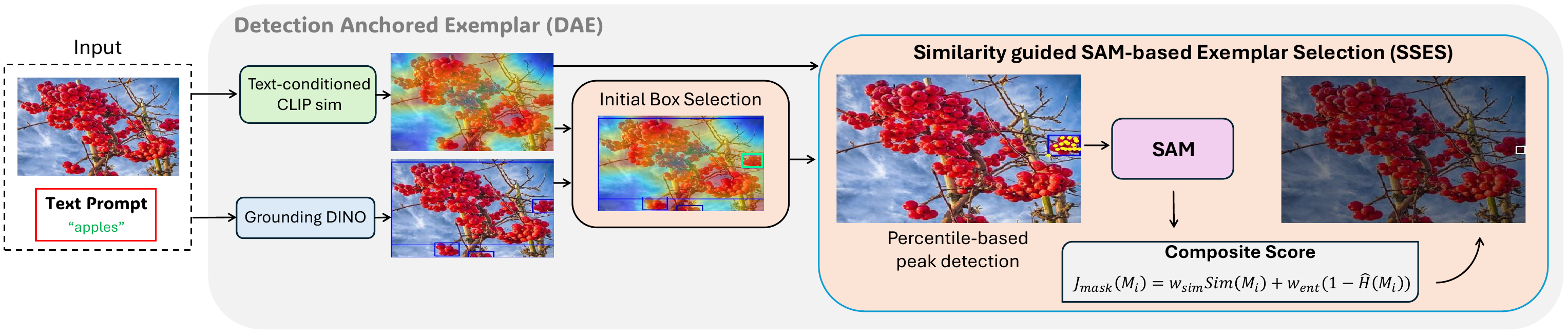}
    \caption{Overview of SSES in the DAE stage. Starting from a detector-derived coarse box (green), CLIP similarity maps are used to extract high-confidence peaks. These peaks guide SAM to generate candidate masks, which are scored based on semantic alignment and spatial consistency. The highest-scoring mask is selected as the refined single-instance exemplar.}
    \label{fig:kpeaks}
\end{figure*}

\begin{figure*}[t]
    \centering
    \includegraphics[width=1\linewidth]{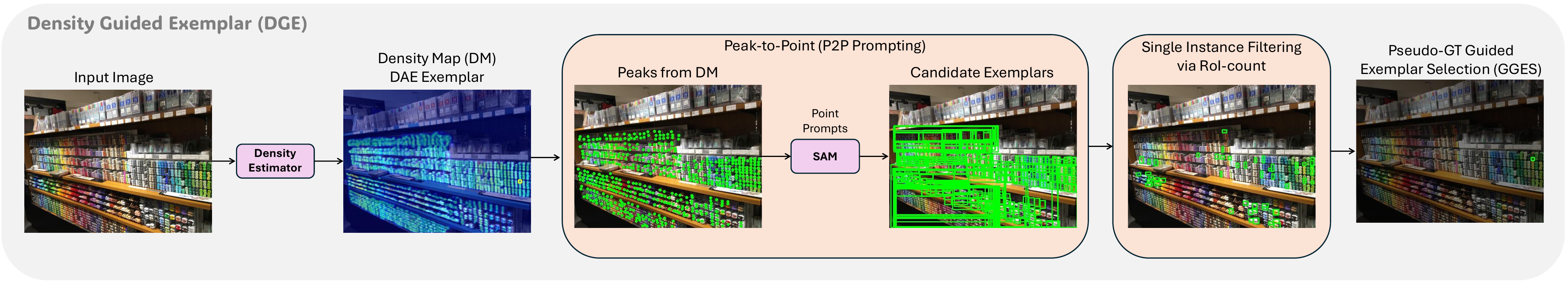}
    \caption{Overview of P2P prompting and single-instance filtering. Density peaks extracted from the density map serve as point prompts to SAM, whose masks are filtered via RoI-count to retain only single-instance regions before being passed to the GGES module for final selection in the DGE stage.}
    \label{fig:p2p_sif}
\end{figure*}

\section{Additional Details}
\label{appendix}

\subsection{Visualization of the SSES Module}
\label{SSES_vis}

Similarity-guided SAM-based Exemplar Selection (SSES), used in the Detection-Anchored Exemplar (DAE) stage, refines the initial coarse detection into a clean, single-instance exemplar by jointly leveraging a text-conditioned similarity map and SAM segmentation. Within the coarse detection box, SSES identifies text-aligned subregions by detecting high-confidence peaks in the similarity map. An adaptive percentile-relaxation strategy accommodates variations in similarity strength across images. As illustrated in Fig.~\ref{fig:kpeaks}, each detected peak is used as a point prompt for SAM, yielding a set of candidate masks. These masks are evaluated using a composite score that combines semantic alignment, measured by the mean percentile rank of the similarities within each mask, and spatial purity, measured by normalized entropy. The resulting score favors masks that are both text-consistent and spatially coherent, enabling SSES to isolate a suitable single-instance exemplar from coarse, cluttered, or ambiguous detections.

\begin{table}[t]
\centering
\caption{Sensitivity analysis of descriptor definition in FRES on FSC-147 dataset.}
\vspace{4pt}
\setlength{\tabcolsep}{6pt}
\fontsize{7.5pt}{7.5pt}\selectfont
\begin{tabular}{lcc}
\toprule
\rowcolor{lightgray!30}
\multirow{2}{*}{Box Descriptor Definition} 
& \multicolumn{2}{c}{FSC-147 (in-domain)} \\
\cmidrule(lr){2-3}
\rowcolor{lightgray!30}
& MAE & RMSE \\
\midrule
\rowcolor{goldenbrown!30}
Avg pool + $\ell_2$-norm & 15.77 & 91.40 \\
Max pool + $\ell_2$-norm & 15.89 & 91.63 \\
\bottomrule
\end{tabular}
\label{tab:descriptor_sensitivity}
\end{table}



\subsection{P2P Prompting and Single-Instance Filtering}
\label{p2p_sif}
We present additional qualitative analysis illustrating the P2P prompting process and the impact of single-instance filtering. As shown in Fig.~\ref{fig:p2p_sif}, given the exemplar from the DAE stage, a density map is first computed using the pretrained counter conditioned on this exemplar. Peaks detected within the density map serve as point prompts to SAM, which generates corresponding masks and bounding boxes. These boxes are then filtered using RoI-count to retain only single-instance exemplars. As observed, this filtering refines the pool of candidate exemplars by discarding multi-instance or noisy regions. The resulting single-instance boxes are subsequently passed to the Pseudo-GT Guided Exemplar Selection (GGES) module to identify the final DGE exemplar.

\begin{table}[t]
\centering
\caption{ \textbf{Left:} Sensitivity to weight value $\alpha$. \textbf{Right:} Effect of number of clusters in FRES.}

\vspace{4pt}

\begin{minipage}{0.50\linewidth}
\centering
\setlength{\tabcolsep}{1.5pt}
\fontsize{7pt}{7pt}\selectfont
\begin{tabular}{lcc|cc}
\toprule
\rowcolor{lightgray!30}
\multirow{2}{*}{$\alpha$} 
& \multicolumn{2}{c}{In-domain} 
& \multicolumn{2}{c}{Cross-domain} \\
\cmidrule(lr){2-3}\cmidrule(lr){4-5}
\rowcolor{lightgray!30}
& MAE & RMSE & MAE & RMSE \\
\midrule
0.25 & 16.36 & 92.22 & 21.48 & 110.96\\
\rowcolor{goldenbrown!30}
0.5  & 15.77 & 91.40 & 21.09 & 110.14 \\
0.75 & 16.18 & 92.35 & 21.44 & 111.15 \\
1    & 17.82 & 94.61 & 22.97 & 113.37 \\
\bottomrule
\end{tabular}
\end{minipage}
\hfill
\begin{minipage}{0.45\linewidth}
\centering
\setlength{\tabcolsep}{1.5pt}
\fontsize{7pt}{7pt}\selectfont
\begin{tabular}{lcc}
\toprule
\rowcolor{lightgray!30}
\multirow{2}{*}{Clusters} 
& \multicolumn{2}{c}{FSC-147} \\
\cmidrule(lr){2-3}
\rowcolor{lightgray!30}
& MAE & RMSE \\
\midrule
\rowcolor{goldenbrown!30}
2 & 15.77 & 91.40 \\
3 & 16.33 & 92.43 \\
4 & 16.81 & 93.26 \\
5 & 17.55 & 93.41 \\
\bottomrule
\end{tabular}
\end{minipage}

\label{tab:combined_alpha_clusters_variants}
\vspace{-6pt}
\end{table}

\subsection{Sensitivity to Design Choices in CountZES}
\label{design_choices_appendix}

\noindent\textbf{Sensitivity to descriptor definition in FRES.}
Each candidate box in FRES is represented using features obtained via spatial average pooling over the SAM feature map within the region, followed by $\ell_2$-normalization. We adopt average pooling as a standard and widely used choice for constructing region-level descriptors, as it provides a stable summary of the overall appearance. We additionally evaluated an alternative descriptor using max pooling. We observe that this results in only a marginal change in performance (see Table~\ref{tab:descriptor_sensitivity}). This behavior is expected in our setting. First, the candidate set passed to FCE is already strongly filtered (via P2P prompting and single-instance filtering), such that most remaining regions correspond to valid object instances with relatively consistent features. Second, SAM features are semantically rich and spatially smooth, causing both average and max pooling to capture similar high-level information. Finally, since FCE is designed to select a consensus exemplar, average pooling is naturally aligned with this objective, whereas max pooling emphasizes localized activations that are less critical at this stage.

\noindent\textbf{Sensitivity to weight parameter $\alpha$ in composite scoring.} The weight $\alpha=0.5$ used in Eqs.~(1) and (4) of the main paper is a fixed prior and not tuned on any dataset. It is chosen as a natural default to equally balance complementary terms (e.g., confidence vs.~entropy, or count consistency vs.~entropy), which are normalized to comparable scales. Empirically, the method is robust to $\alpha$; varying it within $[0.25, 0.75]$ yields only moderate changes in MAE and RMSE on FSC-147 under both in-domain and cross-domain settings (Table~\ref{tab:combined_alpha_clusters_variants}, left). In contrast, setting $\alpha = 1$ (removing the entropy term entirely) significantly degrades performance, highlighting the importance of entropy-based regularization in suppressing noisy or multi-instance regions. Overall, the result demonstrates that the gains stem from the complementary design rather than precise tuning of $\alpha$.

\noindent\textbf{Role of binary clustering in FRES.} FRES operates after P2P prompting and single-instance filtering, where the candidate pool is already restricted to same-class, single-instance regions. In this regime, binary clustering primarily serves as an outlier suppression mechanism, separating a dominant coherent group from noisy or less representative candidates (e.g., partial masks or background spillover), rather than modeling multiple semantic modes. Increasing the number of clusters over-partitions the limited candidate set, leading to deteriorated performance due to unstable centroid estimation and higher sensitivity to fragmentation (Table~\ref{tab:combined_alpha_clusters_variants}, right).

\begin{figure*}[h]
    \centering
    \includegraphics[width=1\linewidth]{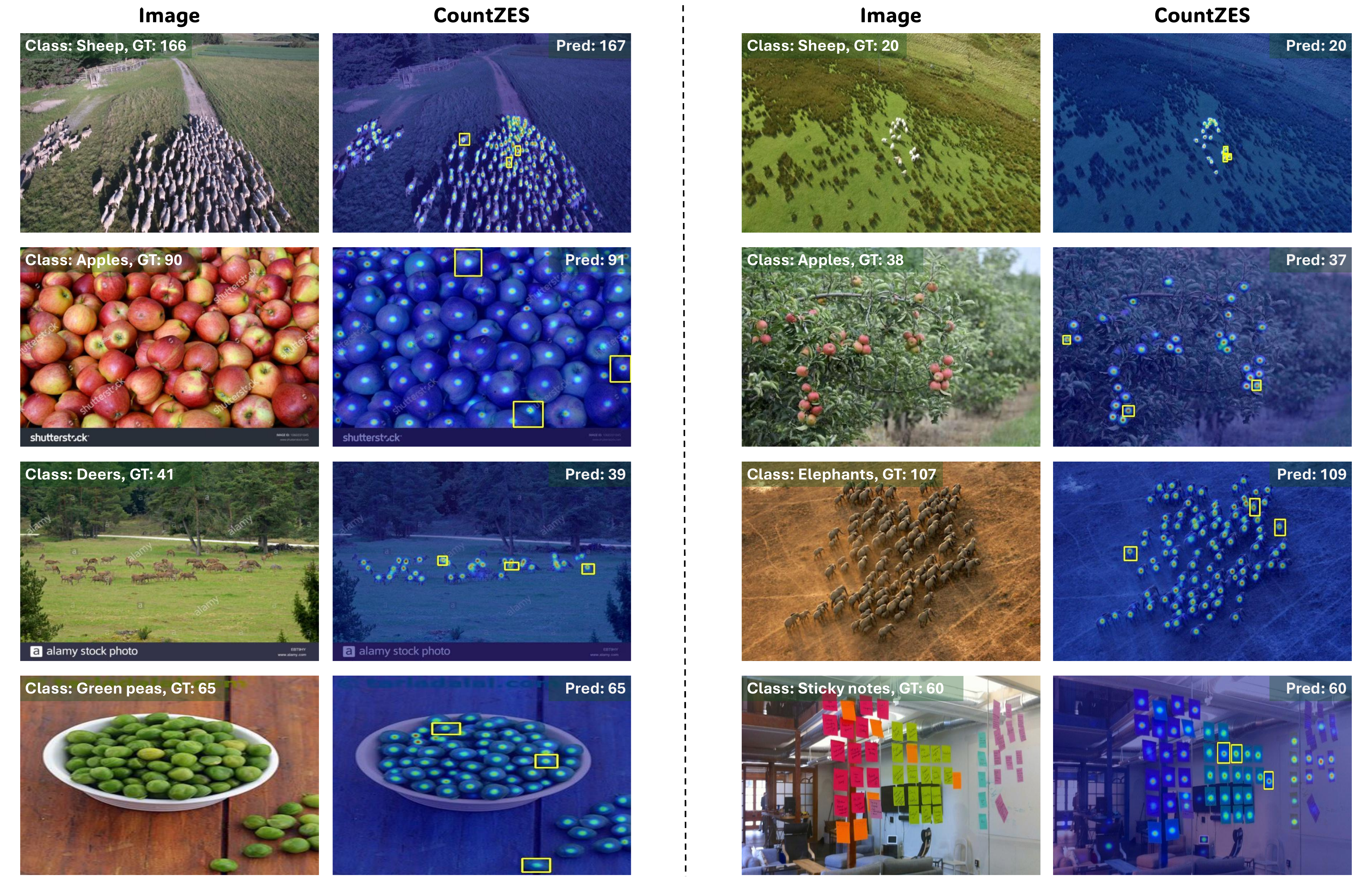}
    \caption{Additional qualitative results of CountZES on FSC-147 dataset.}
    \label{fig:qual_supp}
\end{figure*}

\begin{figure*}[t]
    \centering
    \includegraphics[width=1\linewidth]{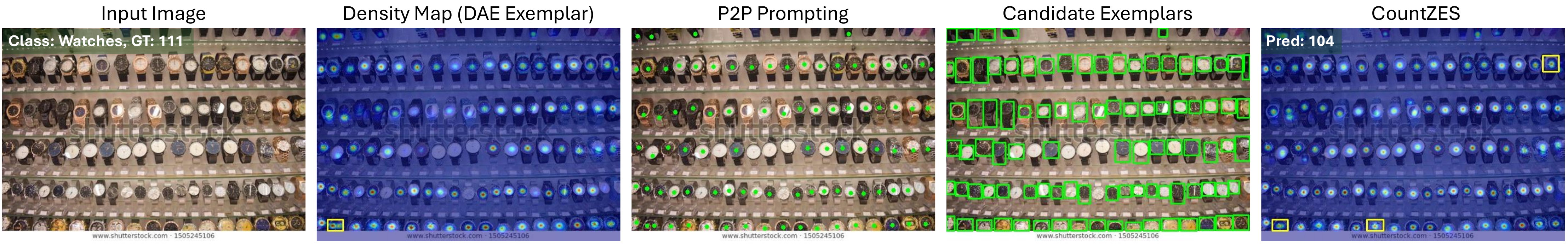}
    \caption{Failure case illustrating selection of incomplete exemplar boxes for the class “watches.” Due to SAM’s limited semantic awareness, point prompts from density peaks segment only the watch dial, excluding the wristband, which produces partial masks and incomplete boxes that lead to suboptimal exemplar selection.}
    \label{fig:limitations1}
\end{figure*}

\begin{figure*}[t]
    \centering
    \includegraphics[width=0.97\linewidth]{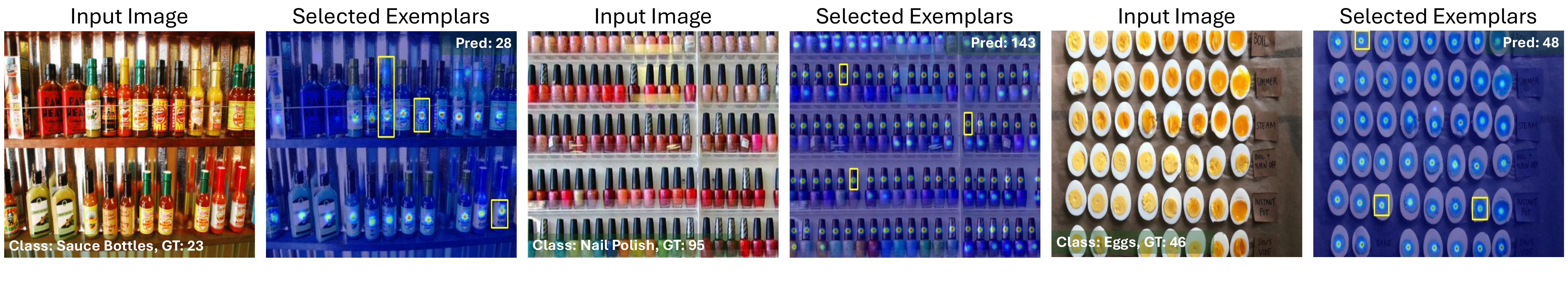}
    \caption{Additional failure cases highlighting CountZES selection of incomplete boxes as exemplars. For “nail polishes,” SAM-based point prompt segmentation in CountZES captures only partial objects; for “comic books,” point prompts isolate characters or cover details instead of the full book; for “eggs,” SAM segments only the yolk region; and for “sauce bottles,” it segments only the label instead of the whole bottle. These examples show how imperfect masks can impair downstream exemplar selection in CountZES.}
    \label{fig:lim}
\end{figure*}

\subsection{Mathematical Details}
\label{maths}

\noindent\textbf{Entropy Computation.} 
For each candidate detection box $\mathbf{b}_i = (x_0, y_0, x_1, y_1)$, we extract the CLIP similarity values from the similarity map $S \in [0,1]^{H \times W}$ within the box region, forming a set $\mathcal{V}_i = \{S(x,y) \mid (x,y) \in \mathbf{b}_i\}$. To estimate the distribution of similarity responses, we construct a histogram with $B=20$ equally spaced bins over the range $[0,1]$, yielding an empirical probability mass function $h_i(k)$ for $k = 1,\dots,B$. We compute the Shannon entropy of this distribution as
\begin{equation}
H(\mathbf{b}_i) = -\sum_{k=1}^{B} \tilde{h}_i(k)\log \tilde{h}_i(k),
\end{equation}
where $\tilde{h}_i(k) = \frac{h_i(k) + \epsilon}{\sum_{j=1}^{B} (h_i(j) + \epsilon)}$ applies Laplace smoothing with $\epsilon=10^{-8}$ for numerical stability. Intuitively, low entropy indicates a concentrated similarity-response distribution, whereas high entropy suggests heterogeneous or noisy responses within the candidate region. To enable fair comparison across candidate boxes, we further perform min--max normalization over all entropy values:
\begin{equation}
\widehat{H}(\mathbf{b}_i) = \frac{H(\mathbf{b}_i) - H_{\min}}{H_{\max} - H_{\min} + \epsilon},
\end{equation}
where $H_{\min}$ and $H_{\max}$ denote the minimum and maximum entropy among valid detections, respectively.

\noindent\textbf{Pseudo-GT Count Estimation.}
Let $\mathcal{B}_{\text{single}} = \{\mathbf{b}_i\}_{i=1}^{N}$ denote the set of candidate single-instance boxes obtained after single instance filtering via density-based RoI count. For each candidate box $\mathbf{b}_i$, we obtain a local count estimate by leveraging the pretrained off-the-shelf density estimator $f_{\text{cnt}}(\cdot)$:
\begin{equation}
\hat{c}_i = f_{\text{cnt}}(I, \mathbf{b}_i), \quad i=1,\dots,N,
\end{equation}
where $\hat{c}_i \in \mathbb{Z}$ represents the predicted number of instances within $\mathbf{b}_i$. These per-box predictions are noisy due to imperfect localization and counter uncertainty. To robustly aggregate these estimates, we model their empirical distribution using kernel density estimation (KDE), yielding a continuous probability density function:
\begin{equation}
\hat{p}(c) = \frac{1}{N h}\sum_{i=1}^{N} K\!\left(\frac{c - \hat{c}_i}{h}\right),
\end{equation}
where $K(\cdot)$ is a Gaussian kernel and $h$ denotes the bandwidth parameter. The pseudo ground-truth count is then defined as the mode of this distribution:
\begin{equation}
c_{\text{pseudo}} = \arg\max_{c}\, \hat{p}(c).
\end{equation}

\noindent In cases where the distribution exhibits negligible variance or insufficient distinct values, we adopt a robust fallback by assigning $c_{\text{pseudo}}$ to the common value if all $\hat{c}_i$ are identical, or to the rounded mean $\lfloor \frac{1}{N}\sum_{i=1}^{N}\hat{c}_i \rceil$ otherwise. This aggregation strategy consolidates multiple uncertain predictions into a stable reference count while mitigating the influence of outliers. The KDE pseudo-GT estimate depends on the bandwidth $h$, which controls the smoothness of the estimated distribution over the candidate counts $\{\hat{c}_i\}$. In our implementation, $h$ is not manually tuned but is computed using the standard data-driven Scott's rule~\cite{scott1992multivariate}, which provides a stable bandwidth without manual tuning. Specifically, the bandwidth $h$ is given by:

\[
h = \sigma \cdot N^{-\frac{1}{d+4}},
\]

\noindent where $N$ is the number of samples, $\sigma$ is the standard deviation of the data, and $d$ is the dimensionality. Since our KDE operates on one-dimensional count values ($d=1$), this simplifies to:

\[
h = \sigma \cdot N^{-\frac{1}{5}}.
\]

\noindent Thus, the bandwidth adapts automatically to both the spread ($\sigma$) and the number of candidate boxes ($N = B_{\text{single}}$), eliminating the need for manual tuning. Importantly, CountZES is not highly sensitive to the exact choice of $h$. The role of KDE is to provide a smoothed consensus over noisy per-box count estimates and identify the dominant mode, rather than recover a precise density function. As a result, moderate variations in $h$ primarily affect the smoothness of the distribution but have minimal impact on the location of the dominant mode, which is used as the pseudo-GT.

For low-density images, where $B_{\text{single}}$ may contain only a few candidates (small $N$), Scott's rule naturally increases the bandwidth ($h \propto N^{-1/5}$), resulting in stronger smoothing. Consequently, it inherently regularizes the KDE in low-density scenarios, preventing overfitting and ensuring that the pseudo-GT is determined by the overall trend of the available candidates rather than spurious fluctuations arising from a small number of samples.



\subsection{Additional Qualitative Results}
\label{qualitative_failure}
Fig.~\ref{fig:qual_supp} presents additional qualitative results of CountZES on the FSC-147 dataset, showcasing its exemplar selection and counting performance across varied object densities and appearances. Fig.~\ref{fig:limitations1} and~\ref{fig:lim} present failure cases.

\end{document}